\documentclass[letterpaper]{article} 
\usepackage{aaai24}  
\usepackage{times}  
\usepackage{helvet}  
\usepackage{courier}  
\usepackage[hyphens]{url}  
\usepackage{graphicx} 
\usepackage{natbib}  
\usepackage{caption} 
\usepackage{algorithm}

\usepackage{multirow}
\usepackage{algpseudocode}
\usepackage{amsmath}
\usepackage{url}
\usepackage{soul}

\usepackage{graphicx}
\usepackage{amssymb}
\usepackage{booktabs}
\usepackage[percent]{overpic}

\usepackage[dvipsnames,table,xcdraw]{xcolor}

\usepackage{makecell}

\usepackage[capitalize]{cleveref}

\usepackage{subcaption}

\usepackage{enumitem}

\usepackage{newfloat}
\usepackage{listings}
\DeclareCaptionStyle{ruled}{labelfont=normalfont,labelsep=colon,strut=off} 
\floatstyle{ruled}
\newfloat{listing}{tb}{lst}{}
\floatname{listing}{Listing}
\title{Hyperspectral Image Reconstruction via Combinatorial Embedding of Cross-Channel Spatio-Spectral Clues}
\author {
    Xingxing Yang\textsuperscript{\rm 1}, 
    Jie Chen\textsuperscript{\rm 1}\thanks{Corresponding Author.}, 
    Zaifeng Yang\textsuperscript{\rm 2}
}
\affiliations {
    \textsuperscript{\rm 1}Department of Computer Science, Hong Kong Baptist University\\
    \textsuperscript{\rm 2}Institute of High Performance Computing, Agency for Science Technology and Research\\
    csxxyang@comp.hkbu.edu.hk, chenjie@comp.hkbu.edu.hk, yang\_zaifeng@ihpc.a-star.edu.sg
}

\usepackage{bibentry}

\begin{document}

\maketitle

\begin{abstract}

Existing learning-based hyperspectral reconstruction methods show limitations in fully exploiting the information among the hyperspectral bands.
As such, we propose to investigate the chromatic inter-dependencies in their respective hyperspectral embedding space. These embedded features can be fully exploited by querying the inter-channel correlations in a combinatorial manner, with the unique and complementary information efficiently fused into the final prediction. We found such independent modeling and combinatorial excavation mechanisms are extremely beneficial to uncover marginal spectral features, especially in the long wavelength bands. In addition, we have proposed a spatio-spectral attention block and a spectrum-fusion attention module, which greatly facilitates the excavation and fusion of information at both semantically long-range levels and fine-grained pixel levels across all dimensions. Extensive quantitative and qualitative experiments show that our method (dubbed CESST) achieves SOTA performance. Code for this project is at: https://github.com/AlexYangxx/CESST.
\end{abstract}

\section{Introduction}
\label{sec:intro}

Combining spectroscopy and image processing techniques, the hyperspectral imaging system (HIS) records rich spectral information along long-range-distributed spectral bands as well as spatial information.
In the past few years, HIS has emerged as a powerful tool in remote sensing \cite{yuan2017hyperspectral}, medical image processing \cite{lu2014medical}, agriculture \cite{adao2017hyperspectral}, etc.
Nonetheless, HIS usually requires a long acquisition time and captures images with limited spatial resolution, which constrained its applications, especially in dynamic or real-time scenarios \cite{arad2020ntire}.
To facilitate and promote the applications of HIS, recent studies have explored efficient data captures, \textit{e.g.}, snapshot compressive imaging system that records 3D hyperspectral cube into the 2D measurement \cite{channing2022spectral,cai2022coarse}. 
However, these methods require expensive, bulky equipment and complicated reconstruction processing for high-fidelity 3D hyperspectral cubes.
To this end, an increased interest in the hyperspectral image (HSI) reconstruction from RGB images using deep learning methods has emerged, which shows great potential due to the handy RGB image capture devices and satisfactory HSI reconstruction performance \cite{yan2020reconstruction, cai2022mst++}.

Conventional hyperspectral reconstruction methods are mainly model-based, \textit{e.g.}, sparse coding \cite{Arad2016SparseRO}, which fails in exploring the intrinsic spectral relations between input RGB images and the corresponding hyperspectral images and suffers from representation capacity. 
Over the years, an enormous amount of deep-learning-based research has been developed that mainly focuses on deep convolutional neural networks (CNNs) in an end-to-end manner \cite{wang2018hyperreconnet, wang2019hyperspectral}.
However, CNNs still fail to capture long-range dependencies and rely on delicately designed modules.

\begin{figure}
\begin{overpic}[width=0.481\textwidth]{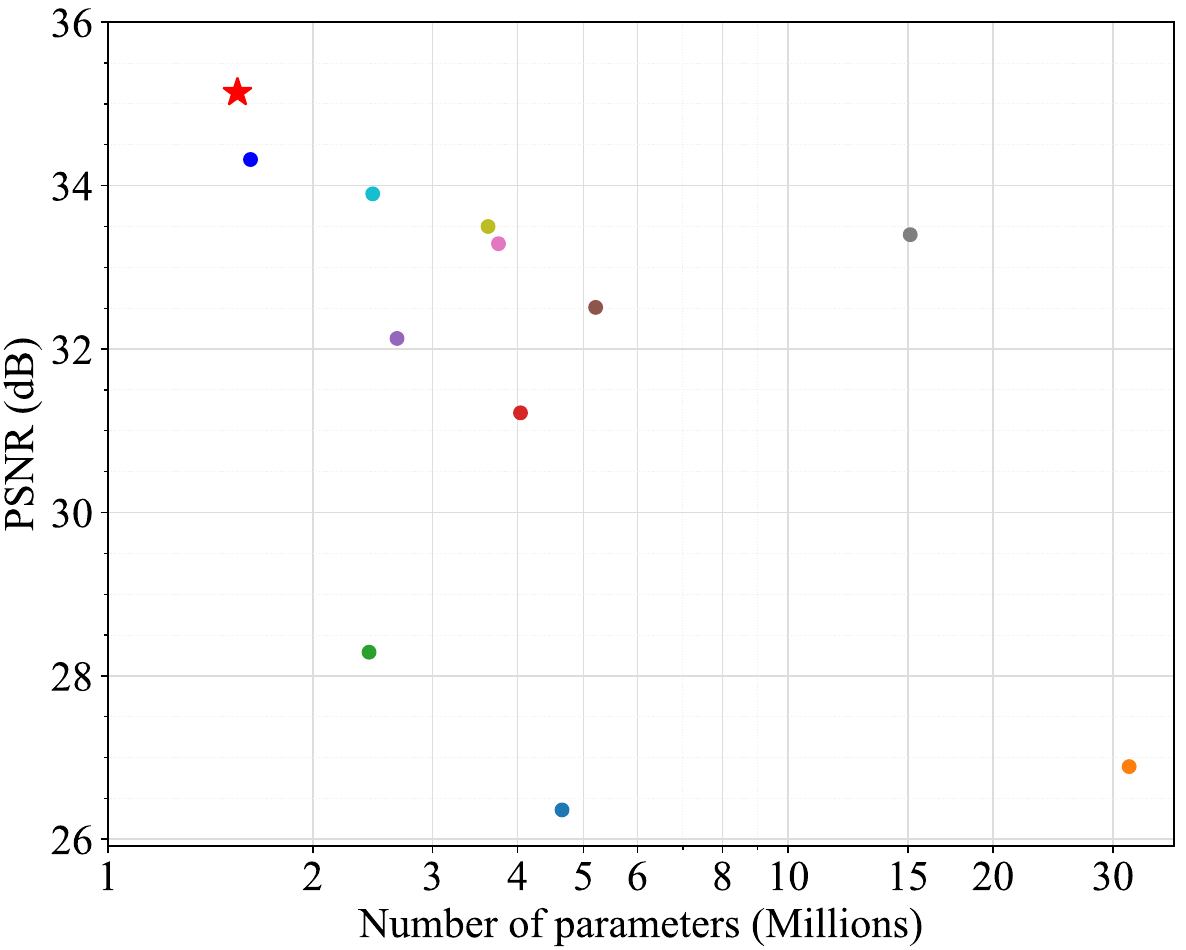}
\put(50,11){\color{NavyBlue}\small HSCNN++\cite{shi2018hscnn+}}
\put(59,18){\color{orange}\small HRNet+\cite{zhao2020hierarchical}}
\put(55,56){\color{gray}\small Restormer\cite{zamir2022restormer}}
\put(25,28){\color{ForestGreen}\small EDSR\cite{lim2017enhanced}}
\put(48,43){\color{red}\small AWAN\cite{li2020adaptive}}
\put(26,47){\color{violet}\small HDNet\cite{hu2022hdnet}}
\put(52,50){\color{brown}\small HINet\cite{chen2021hinet}}
\put(45,59){\color{pink}\small MIRNet\cite{zamir2020learning}}
\put(48,63){\color{olive}\small MPRNet\cite{zamir2021multi}}
\put(33,68){\color{cyan}\small MST-L\cite{cai2022mask}}
\put(11,63){\color{blue}\small MST++\cite{cai2022mst++}}
\put(11,75){\color{red}\small CESST (Our Method)} 
\end{overpic}
\caption{HSI reconstruction on  NTIRE2022 HSI dataset \cite{arad2022ntire}.} 
\label{performance}
\end{figure}

\begin{figure}
  \centering
  \begin{subfigure}{0.43\linewidth}
  \includegraphics[width=0.95\linewidth]{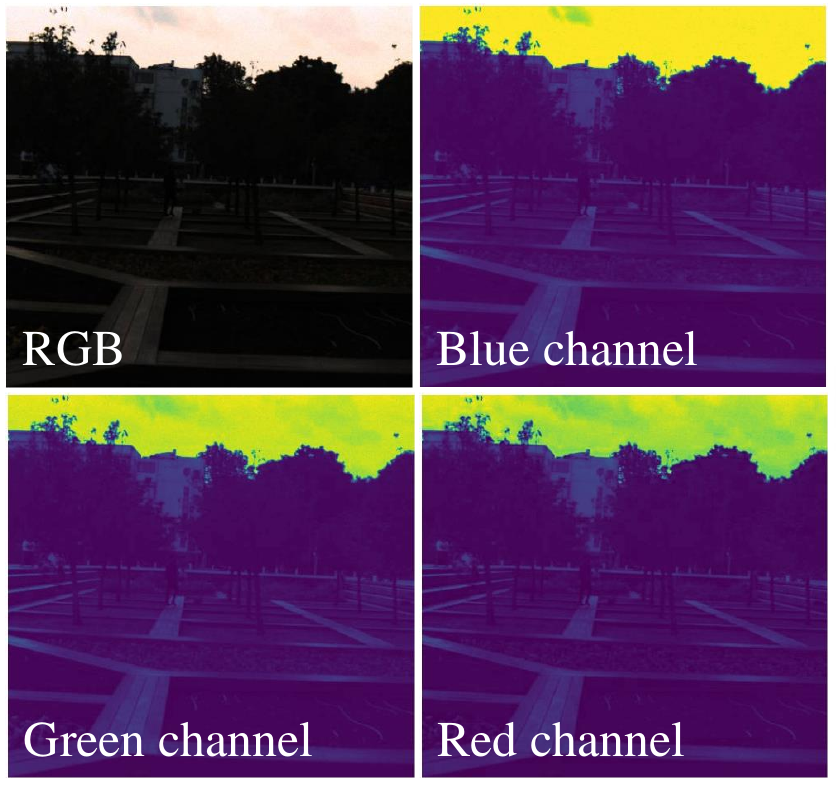}
  \caption{Channel differences.}
    \label{fig:short-a}
  \end{subfigure}
  \hfill
  \begin{subfigure}{0.54\linewidth}
    \includegraphics[width=0.98\linewidth]{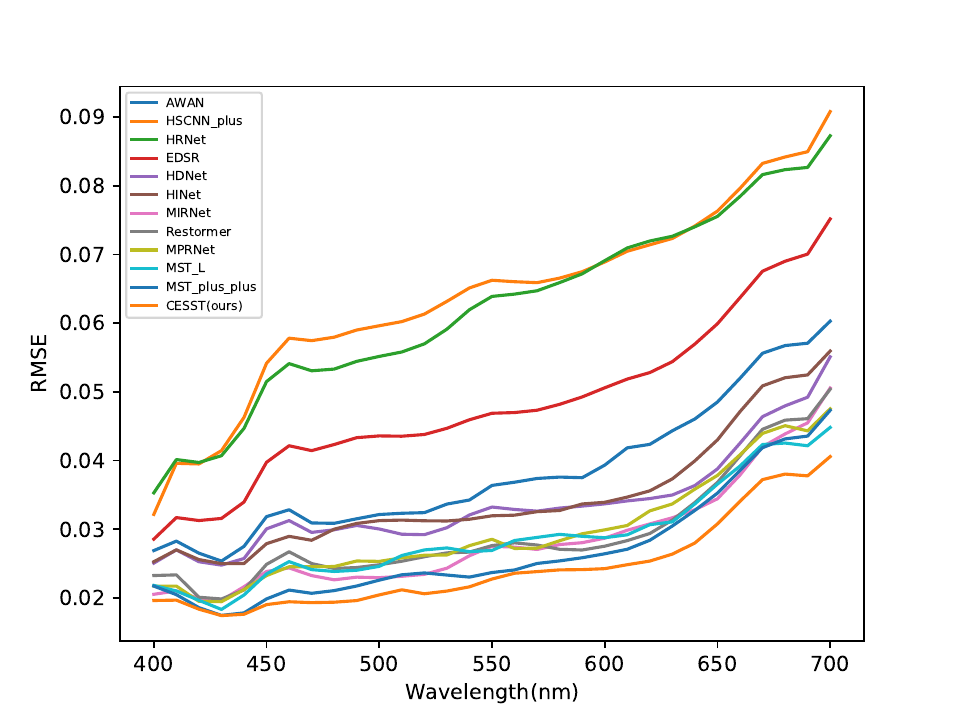}
  \caption{RMSE loss.}
    
    \label{fig:short-b}
  \end{subfigure}
  \caption{Limitation of existing methods. Fig. \ref{fig:short-a} shows an example of visual differences between R, G, and B channels of the same RGB image. As can be seen, the red channel contains more texture energy than the other two channels, and most existing methods show compromised performance over the long-wavelength range. Fig. \ref{fig:short-b} shows the RMSE loss across all bands in the validation set. All existing methods show a dramatic rise of RMSE loss in the long-wavelength bands, while our method slows such deterioration and achieves the lowest RMSE loss across all bands.}
  \label{channel difference}
\end{figure}

\begin{figure*}
  \centering
  \includegraphics[width=0.9\linewidth]{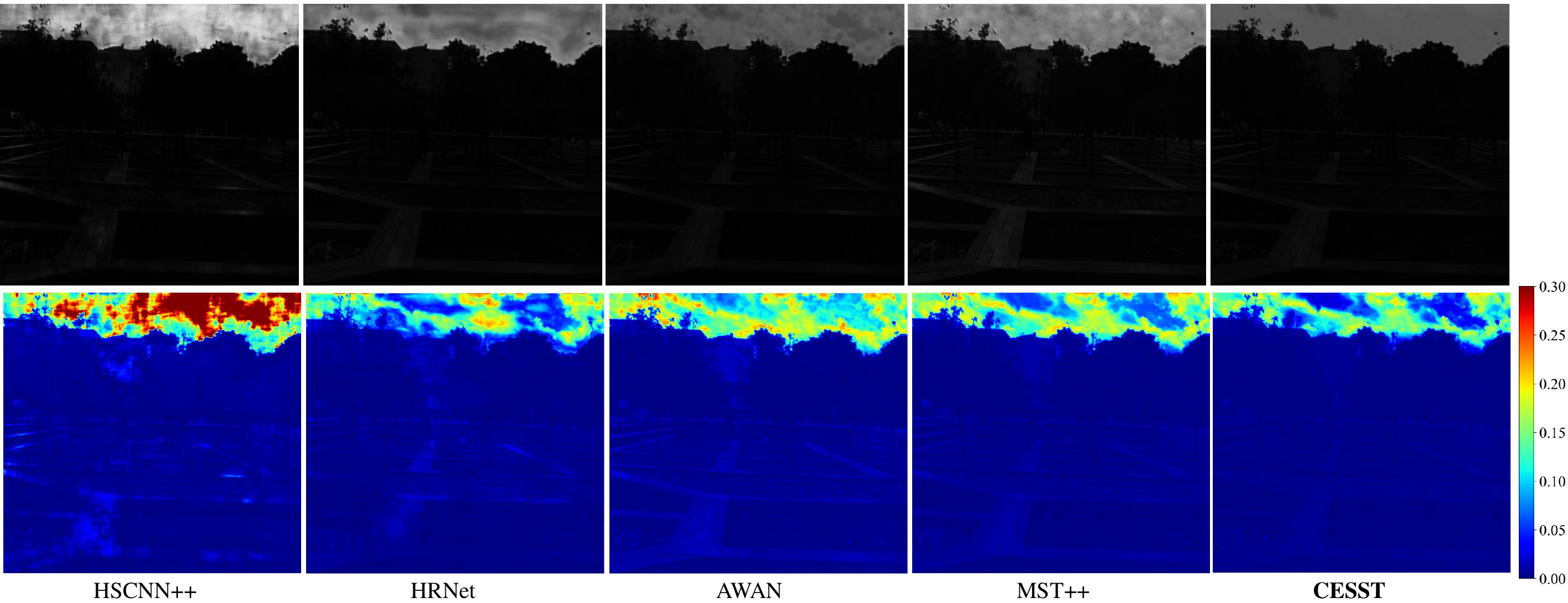}
  \caption{The visual results of the 30th band and the reconstruction error images of an HSI chosen from the validation set of NTIRE2022 HSI dataset \cite{arad2022ntire} predicted by  HSCNN+ \cite{shi2018hscnn+}, HRNet \cite{zhao2020hierarchical}, AWAN \cite{li2020adaptive}, MST++ \cite{cai2022mst++} and our method (CESST). The error images are the heat maps of RMSE (along spectral direction) between the ground truth and the recovered HSI.}
  \label{visual_compar_3}
\end{figure*}

Recently, vision transformer \cite{liu2021swin} (ViT) has been introduced into computer vision from natural language processing (NLP) and shows great potential in learning long-range dependencies and non-local self-similarities. 
However, existing frameworks for HSI reconstruction have two main limitations: 
(\textbf{i}) the complexity of the standard global transformer \cite{dosovitskiy2020image} is quadratic to the spatial dimension, which occupies substantial computational resources.
(\textbf{ii}) Although the Swin-transformer \cite{liu2021swin} and spectral-wise transformer \cite{cai2022mst++} achieve linear complexity to the spatial dimension via window-based multi-head self-attention (MSA) or calculating spectral-wise self-attention maps, they all focus on one dimension of the 3D hyperspectral cube, \textit{i.e.}, spatial or spectral dimension. Most importantly, as shown in Fig. \ref{channel difference}, \textbf{\textit{all these existing reconstruction methods treat the problem trivially by directly studying the correlations in the hyperspectral space}}. Specifically, these methods brutally combine and project the features (with different energy and noise characteristics) from RGB channels into the high dimensional spectral space in the early stages, which would inevitably sacrifice some critical information from the R, G, or B channels.

In this study, we propose a novel hyperspectral image reconstruction framework that excavates the unique and complementary information among the RGB input channels in a \underline{\textbf{C}}ombinatorial manner for efficient \underline{\textbf{E}}mbedding of \underline{\textbf{S}}patio-\underline{\textbf{S}}pectral clues based on a \underline{\textbf{T}}ransformer structure (CESST), which achieves the best PSNR-Params performance compared with SOTA methods in Fig. \ref{performance}, and a typical visual comparison is provided in Fig. \ref{visual_compar_3}. 
The novelty and technical contributions are generalized as follows:
\begin{itemize}[noitemsep]
\item We propose a novel framework for hyperspectral image reconstruction, which first fully excavates the intra-channel spatio-spectral features in the projected high-dimensional embedding space before inter-channel fusion. Such channel-wise independent modeling procedure ensures unique local spectral features are well uncovered and preserved;
\item We propose a novel Spectrum-Fusion Attention Module (SFAM) that exhaustively queries and explores cross-channel correlations in a combinatorial manner via six parallel transformer branches. SFAM fully excavates complementary information for comprehensive inter-channel fusion;
\item An efficient plug-and-play spatio-spectral attention block (SSAB) is designed to simultaneously extract spatio-spectral features at both semantically long-range levels and fine-grained pixel levels across all dimensions, while keeping the complexity linear to the spatial dimension;
\item Both quantitative and qualitative experiments demonstrate that our CESST framework significantly outperforms SOTA methods while requiring fewer Parmas. 
\end{itemize}

\section{Related Work}
\label{sec: related work}

\subsection{Hyperspectral Image Reconstruction}

Two basic approaches are currently being adopted in research into HSI reconstruction from RGB images: model-based and deep learning-based methods. 
Most model-based methods \cite{Arad2016SparseRO, aeschbacher2017defense, jia2017rgb, robles2015single} mainly focus on using hand-crafted priors to conduct spectrum interpolation along the channel dimension. For example, Arad \textit{et al.} \cite{Arad2016SparseRO} proposed to address this issue by leveraging hyperspectral prior to creating a sparse dictionary of HSIs and their corresponding RGB projections. Robles \textit{et al.} \cite{robles2015single} further proposed to employ color and texture information to assist the reconstruction process subject to the material properties of the objects in the scene. However, these model-based methods rely heavily on hand-crafted priors and suffer from poor representation capacities. Meanwhile, they do not take the spatial context into consideration. 

Recently, inspired by the rapid progress of deep learning in image restoration \cite{tu2022maxim, helminger2021generic, chen2022attention, shao2020domain}, CNNs have been widely exploited to learn the implicit mapping relation from RGB to HSI \cite{wang2018hyperreconnet, wang2019hyperspectral, shi2018hscnn+, fubara2020rgb}, which learn the spatial contextual information in a statistic sense. For instance, HSCNN \cite{xiong2017hscnn} proposed to upsample the input RGB image along the spectral dimension and learn the corresponding enhanced HSI using deep residual convolutional blocks. Alvarez-Gila \textit{et al.} \cite{alvarez2017adversarial} further proposed a conditional generative adversarial framework to deal with the paired-data insufficiency. Many other methods have produced impressive results by designing delicate architectures, including UNet \cite{can2018efficient}, Resnet \cite{stiebel2018reconstructing}, and self-attention mechanisms \cite{wang2020dnu}. However, these CNN-based methods show limitations in capturing long-range inter-dependencies and non-local self-similarities.

\begin{figure*}[t]
    \centering
  { 
      \includegraphics[width=1\linewidth]{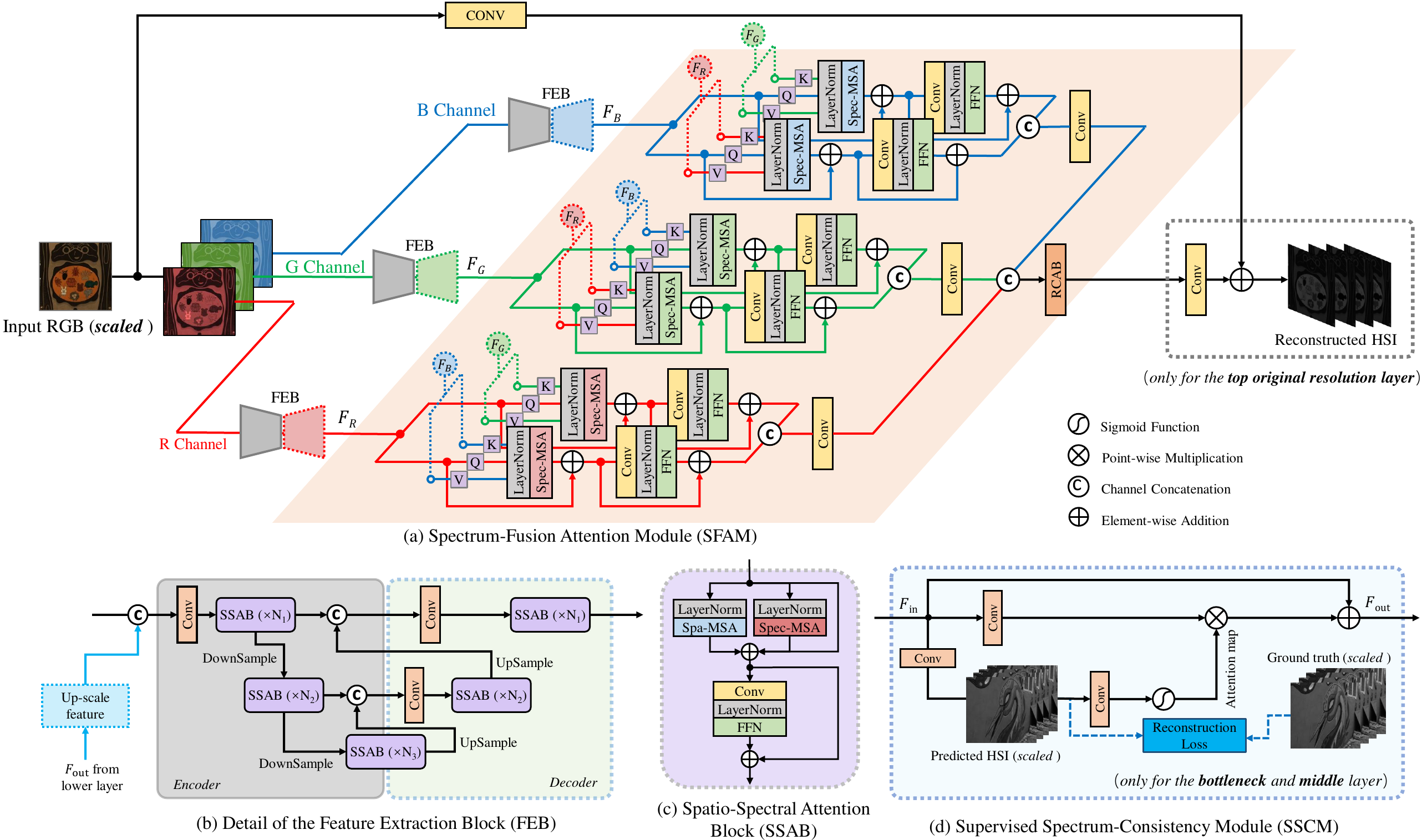}}
  \caption{Illustration of the proposed CESST framework, which consists of three layers of similar structures as shown on the top row, which represents the top original resolution layer. A middle layer and a bottleneck layer share similar structures but with slight differences.}
    \label{CESST} 
\end{figure*} 

\subsection{Vision Transformer}
Since the vision Transformer (ViT) \cite{dosovitskiy2020image} was first introduced into vision tasks, there has been a wave of enthusiasm due to its strength in capturing long-range correlations between spatial contexts. Since the complexity of standard global transformers \cite{chen2021pre, dosovitskiy2020image} is quadratic to the spatial dimension of input images, many researchers introduce the local-processing idea of CNNs into transformer blocks \cite{liu2021swin, zhou2021swin} to reduce the computational burden. For instance, Liu \textit{et al.} proposed to leverage local window-based MSA, whose computational complexity is linear to the spatial dimension. Cai \textit{et al.} \cite{cai2022mask} further proposed a spectral-wise MSA to calculate the self-attention map along the channel dimension for HSI reconstruction. Nonetheless, neither spatial window-based MSA nor spectral-wise MSA considers spatial and spectral information, limiting the representation capacity.


\section{The Proposed Method}
\label{sec: method}

\textbf{Motivation.} Existing frameworks \cite{cai2022mst++, wang2019hyperspectral, shi2018hscnn+, fubara2020rgb} project the RGB image directly to the high-dimensional hyperspectral space in an early stage. Such brutal transformation sacrifices potentially crucial intra-channel features, as shown in Fig. \ref{channel difference}, and it would be more difficult to learn from the inter-channel correlations in subsequent stages(the failure case corresponding to Fig. \ref{channel difference} is illustrated in Fig. \ref{visual_compar_3}). 
As such, we propose to first fully excavate the intra-channel spatio-spectral features in the projected high-dimensional embedding space before inter-channel fusion, ensuring local spectral features are well uncovered and preserved.
\subsection{Network Architecture}

We propose a multi-scale encoder-decoder architecture for HSI reconstruction, which has three layers of similar structures as shown on the top row of Fig. \ref{CESST}, with each layer focusing on different scales (full, half, and quarter sizes). 

At each scale, three encoder-decoder feature extraction blocks (FEBs) are designed to learn the contextual features of each channel independently (\textit{e.g.}, R, G, or B).
Unlike the other methods that brutally combine and project the RGB channels into the high-dimensional spectral space in the early stages, our channel-wise independent modeling procedure ensures unique local spectral features are well uncovered and preserved. 
FEBs adopt UNet \cite{yang2021attention} as the backbone to extract both contextual and spectral features crucial for spectrum reconstruction. Specifically, as shown in Fig. \ref{CESST}~(b), each FEB comprises two encoder blocks, one bottleneck block, and two decoder blocks. Each block consists of a spatio-spectral attention block (SSAB), shown in Fig. \ref{CESST}~(c). 
Subsequently, a spectrum-fusion attention module (SFAM), illustrated in Fig. \ref{CESST}~(a), is cascaded to the output of the three FEBs. 
The SFAM exhaustively queries and explores cross-channel correlations in a combinatorial manner via six parallel transformer branches and comprehensively fuses the complementary information from R, G, and B channels.
Finally, inspired by \cite{zamir2021multi}, a supervised spectrum-consistency module (SCCM) is cascaded to generate spectrum-consistent predictions as well as cross-scale features with the supervision of ground-truth HSI signals (as shown in Fig. \ref{CESST} (d)), which is only embedded in the middle and bottleneck branches. 
The cross-scale features provide informative guidance from a lower scale to a larger scale, thus formulating a coarse-to-fine reconstruction (indicated by the light blue lines in Fig. \ref{CESST}).

\subsection{Spatio-Spectral Attention Block}
\label{SSAB}

HSI contains plentiful spatio-spectral clues; however, existing CNN-based feature extraction blocks struggle to model the non-local self-similarities. Meanwhile, transformer-based blocks only take one-dimensional features into account (\textit{i.e.}, spatial \cite{liu2021swin} or spectral \cite{cai2022mst++}). 
To address these issues, we propose a dual-dimensional transformer-based feature extraction block embedded in the FEB, denoted as spatio-spectral attention block (SSAB), to extract both spatial and spectral features and increase the learning capacity.
As shown in Fig. \ref{CESST}~(c), the SSAB consists of a parallel spatial-MSA and spectral-MSA, which calculates both the spatial multi-head self-attention and the spectral multi-head self-attention in parallel, and then feeds both features to enhance cross-dimensional interaction.

\begin{figure}[t]
    \centering
  { 
      \includegraphics[width=0.95\linewidth]{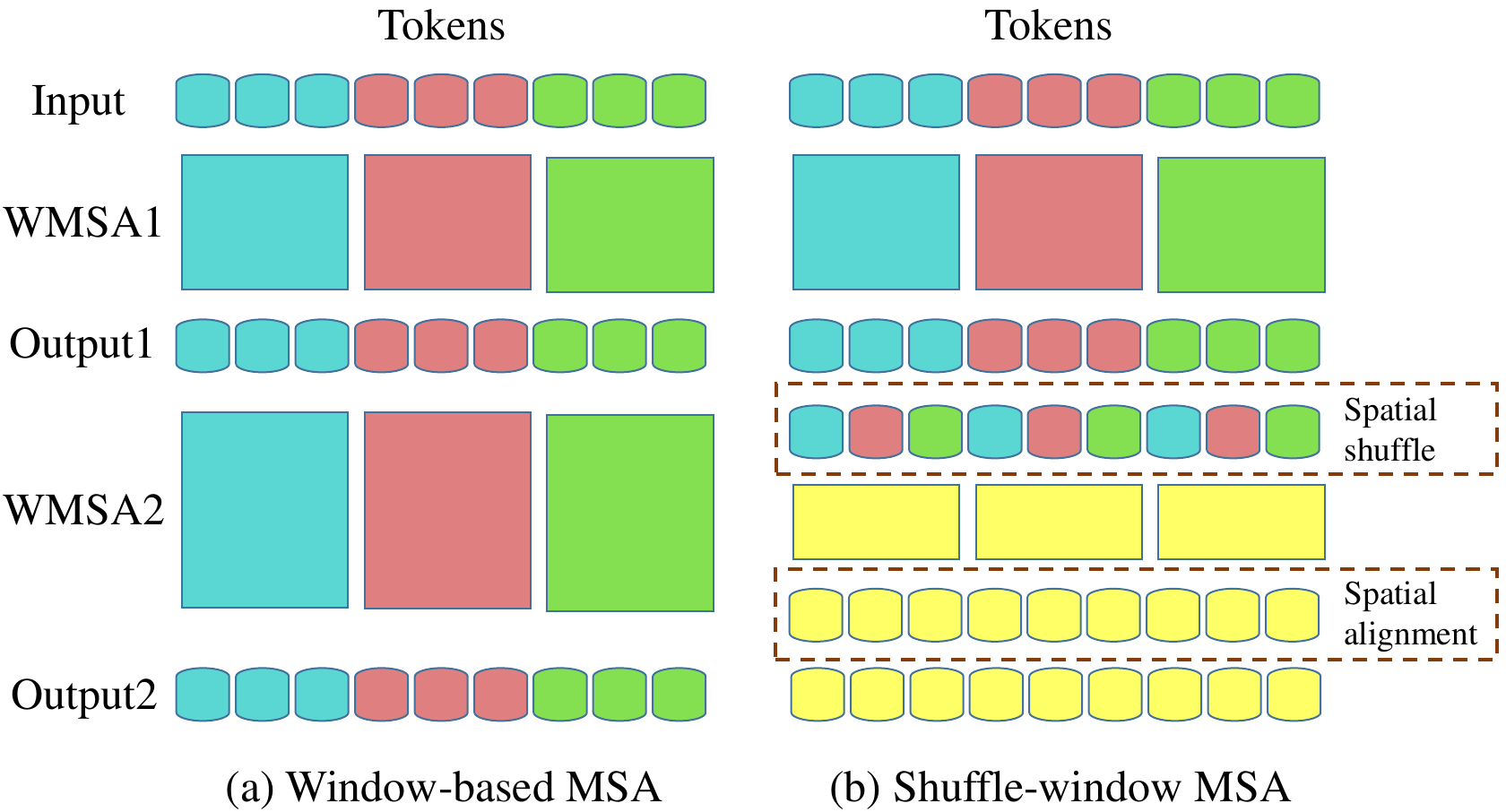}}
  \caption{Comparison of traditional window-based MSA and our shuffle-window MSA. WMSA represents window-based multi-head self-attention. (a) two stacked window-based transformer blocks, where each output token only relates to tokens within the same window, without any cross-window interaction; (b) WMSA2 takes data from different windows after WMSA1 by spatial shuffle and alignment, which introduces global cross-window interaction.}
    \label{shuffle-MSA} 
\end{figure}

Note that our proposed spatial-MSA is different from conventional window-based MSA \cite{liu2021swin}, which suffers from limited receptive fields within non-overlapping position-specific windows. 
As shown in Fig. \ref{shuffle-MSA}, our spatial-MSA consists of a normal window-based MSA, followed by a shuffle-window MSA, to build long-range cross-window interactions. The main difference between conventional window-based MSA and shuffle-window MSA is the spatial shuffle mechanism. 
To be specific, we assume a Window-based MSA with window size M whose input has N tokens; we first reshape the output spatial dimension into $(M, N/M)$, transpose and then flatten it back as the input of the next layer. 
This operation puts the tokens from distant windows together and helps build long-range cross-window connections. 
Note that spatial shuffle requires the spatial alignment operation to adjust the spatial tokens into the original positions for spatially aligning features and image content. 
The spatial alignment operation first reshapes the output spatial dimension into $(N/M, M)$, transposes it, and then flattens it, which is an inverse process of the spatial shuffle. 
Moreover, considering that the "grid issue" widely exists when using window-based transformers to deal with high-resolution images, we introduce a depth-wise convolution layer between the normal window-based MSA and shuffle-window MSA via a residual connection.  The kernel size of the convolution layer is the same as the window size.
On the other hand, the spectral-MSA is mostly inspired by \cite{zamir2022restormer, cai2022mst++}, which treats the spectral feature map as a token and thus focuses on more non-local spectral self-similarities. 

\begin{table}[t]
\small
\caption{Comparison of the properties of different MSAs.}
\vspace{-6pt}
\centering
\label{MSA property} 
\resizebox{1\columnwidth}{!}{
\begin{tabular}{lcccc}
\toprule
MSA scheme   &  Receptive Field & Complexity to HW & Calculating Wise \\ 
\toprule
\textit{ViT \cite{dosovitskiy2020image}}~       & Global & Quadratic & Spatial \\
\textit{Swin-T \cite{liu2021swin}}~             & local & Linear & Spatial \\
\textit{Restormer \cite{zamir2022restormer}}~   & Global & Linear & Spectral \\
\textit{MST++ \cite{cai2022mst++}}~             & Global & Linear & Spectral \\
\textit{\textbf{CESST}}~                        & Global & Linear & Spatio-spectral\\
\bottomrule
\end{tabular}
}
\end{table}

We further summarize the main properties of existing transformer-based blocks, including ViT \cite{dosovitskiy2020image} (global MSA), Swin-Transformer \cite{liu2021swin} (Window-based MSA), Restormer \cite{zamir2022restormer} (spectral MSA), MST++ \cite{cai2022mst++} (spectral MSA), and our CESST (spatio-spectral MSA) in Table \ref{MSA property}. 
Our CESST computes global receptive fields and models both spatial and spectral self-similarities with linear computational costs.

\subsection{Spectrum-Fusion Attention Module}
To further improve the feature utilization and interactivity within the three learned hyperspectral representations (\textit{i.e.}, $F_R$, $F_G$, $F_B$), we design an effective feature fusion module, named spectrum-fusion attention module, including two parts: channel learning and spectrum fusion.

In \textbf{channel learning}, we propose to extract the correlations between each two of the three learned hyperspectral representations and then fuse them to generate the final fine-grained reconstructed HSI. 
As shown in Fig. \ref{CESST} (a), taking $F_R$ as an example, we take the  $F_G$ as the \textit{Value} and \textit{Query}, and  $F_R$ as the \textit{Key}, and then feed them into the spectral-MSA to learn the correlation between $F_R$ and $F_G$.
The learning of the correlation between $F_R$ and $F_B$ is also carried out in a similar way. 
Then, the $F_G$-enriched $\mathcal{F}_{R}^{RG}$ and the $F_B$-enriched $\mathcal{F}_{R}^{RB}$ are concatenated to fed into a convolutional layer to generate $F_G$-$F_B$-enriched $\mathcal{F}_R$, which can be formulated as:
\begin{align}
\label{F_RG}
\mathcal{F}_{R}^{RG} &=F_{\text {S-MSA }}\left(\mathbf{Q}_{F_{R}}, \mathbf{K}_{F_{G}}, \mathbf{V}_{F_{G}} \right) \in \mathbb{R}^{31 \times H \times W},\\
\label{F_RG}
\mathcal{F}_{R}^{RB} &=F_{\text {S-MSA }}\left(\mathbf{Q}_{F_{R}}, \mathbf{K}_{F_{B}}, \mathbf{V}_{F_{B}} \right) \in \mathbb{R}^{31 \times H \times W},\\
\label{F_R}
\mathcal{F}_{R} &=F_{\text {conv }}\left(F_{\text {concat }}\left[\mathcal{F}_{R}^{RG}, \mathcal{F}_{R}^{RB}\right] \right) \in \mathbb{R}^{31 \times H \times W},
\end{align}
where $F_{\text {S-MSA }}(\cdot)$ is spectral-MSA operation (including layernorm, residual connection, and feed-forward operations for simplicity, $F_{\text {conv }}(\cdot)$ is a $3\times3$ convolutional layer, $F_{\text {concat }}(\cdot)$ is a concatenation process.
Similarly, $F_R$-$F_B$-enriched $\mathcal{F}_G$ and $F_R$-$F_G$-enriched $\mathcal{F}_B$ also have such two channel learning branches. Thus, the channel learning part has six branches. 
In \textbf{spectrum fusion}, the three representative hyperspectral features $\mathcal{F}_R$, $\mathcal{F}_G$, $\mathcal{F}_B$ are concatenated first and then fed into a residual coordinate attention block \cite{yang2021attention} (RCAB) to generate fine-grained pixel-level HSI signals, which can be formulated as:
\begin{equation}
\label{HSI}
\mathbf{X}=F_{\text {RCAB }}\left(F_{\text {concat }}\left[\mathcal{F}_{R}, \mathcal{F}_{G}, \mathcal{F}_{B}\right] \right) \in \mathbb{R}^{31 \times H \times W}.
\end{equation}


\subsection{Objective Function}

To supervise reconstructed HSIs at any given scale $s=1,2,..., S$, we employ the $L_{MIX}$ loss \cite{zhao2016loss}, which combines both SSIM loss and L1 loss, as well as the mean relative absolute error (MRAE), to formulate a supervised consistency constraint on both pixel and feature levels:
\begin{align}
\mathcal{L}_{\text{MIX}}& =\sum_{s=1}^{3}\left[(\mathbf{X}^s, \mathbf{Y}^s)_{mix}\right], \\
\mathcal{L}_{\text{MRAE}}& =\sum_{s=1}^{3}\left[\frac{|\mathbf{Y}^s-{\mathbf{X}}^s|}{\mathbf{Y}^s}\right].
\end{align} 
Here $\mathbf{Y}^s$ represents the ground-truth image in each scale.

\textbf{Total Loss.}
The full objective function is expressed as:
\begin{align}
\label{total_losses}
\mathcal{L} & = \mathcal{L}_{\text{MIX}} +\lambda_{1}\mathcal{L}_{\text{MRAE}},
\end{align}
Where $\lambda_{1}$ is the hyperparameter that controls the relative importance of the two loss terms, empirically set to 100. 


\begin{table*}[t]\small
\begin{center}
\caption{Comparison with SOTA methods on  NTIRE2022 HSI dataset. The best results are highlighted in \textcolor{red}{red}.}
  \vspace{-6pt}
\centering
\label{quantitative_2022_1}
\resizebox{2.1\columnwidth}{!}{
\begin{tabular}{lcccccccccccc}
\toprule
\multirow{2}{*}{Method}   & \multicolumn{7}{c}{NTIRE2022 HSI Dataset}  & \multicolumn{5}{c}{ICVL Dataset} \\
& Params(M) & FLOPs(G) & ERGAS & SAM & MRAE & RMSE & PSNR & ERGAS & SAM & MRAE & RMSE & PSNR \\
\toprule
\textit{HSCNN+ \cite{shi2018hscnn+}}~           & 4.65 & 304.45 & 228.80 & 0.1093 & 0.3814 & 0.0588 & 26.36 & 247.31 & 0.1108 &  0.2178 & 0.0533 & 26.28   \\
\textit{HRNet \cite{zhao2020hierarchical}}~     & 31.70 & 163.81 & 107.23 & 0.0855 & 0.3476 & 0.0550 & 26.89 & \textcolor{red}{\textbf{101.02}} & \textcolor{red}{\textbf{0.0806}} &  0.2155 & 0.0497 & 26.93  \\
\textit{AWAN \cite{li2020adaptive}}~            & 4.04 & 270.61 & 147.57 & 0.0960 & 0.2500 & 0.0367 & 31.22 & 169.93 & 0.1054 &  0.1887 & 0.0382 & 30.33  \\
\textit{MST++ \cite{cai2022mst++}}~             & \underline{1.62} & \textcolor{red}{\textbf{23.05}} & 107.23 & \underline{0.0852} & \underline{0.1645} & \underline{0.0248} & \underline{34.32}   & 118.33 & 0.0972 & 0.1776 & \underline{0.0351} & \underline{31.41} \\
\textit{EDSR \cite{lim2017enhanced}}~           & 2.42 & 158.32 & 212.51 & 0.0983 & 0.3277 & 0.0437 & 28.29 &  235.83 & 0.1039 & 0.1972 & 0.0467 & 27.51 \\
\textit{HDNet \cite{hu2022hdnet}}~              & 2.66 & 173.81 & 133.72 & 0.1006 & 0.2048 & 0.0317 & 32.13 & 177.47 & 0.1153 & 0.1942 & 0.0418 & 29.28 \\
\textit{HINet \cite{chen2021hinet}}~            & 5.21 & \underline{31.04} & 140.82 & 0.0937 & 0.2032 & 0.0303 & 32.51  & 152.45 & 0.0983 & \underline{0.1663} & 0.0365 & 30.43   \\
\textit{MIRNet \cite{zamir2020learning}}~       & 3.75 & 42.95 & 115.38 & 0.0944 & 0.1890 & 0.0274 & 33.29   & 131.43 & 0.0998 & 0.1797 & 0.0357 & 30.76    \\
\textit{Restormer \cite{zamir2022restormer}}~   & 15.11 & 93.77 & 112.05 & 0.0983 & 0.1833 & 0.0274 & 33.40 & 130.17 & 0.1003 & 0.1689 & \underline{0.0351} & 31.01 \\
\textit{MPRNet \cite{zamir2021multi}}~          & 3.62 & 101.59 & \underline{101.50} & 0.0901 & 0.1817 & 0.0270 & 33.50 & 131.11 & 0.0979 &  0.2138 & 0.0411 & 29.09 \\
\textit{MST-L \cite{cai2022mask}}~              & 2.45 & 32.07 & 112.57 & 0.0931 & 0.1772 & 0.0256 & 33.90  & 122.36 & \underline{0.0893} & 0.1845 & 0.0380 & 30.75 \\
\textit{\textbf{CESST (ours)}}~                 & \textcolor{red}{\textbf{1.54}} & 90.18 & \textcolor{red}{\textbf{98.74}} & \textcolor{red}{\textbf{0.0791}} & \textcolor{red}{\textbf{0.1497}} & \textcolor{red}{\textbf{0.0225}}  & \textcolor{red}{\textbf{35.14}} & \underline{109.47} & 0.0917 & \textcolor{red}{\textbf{0.1230}} & \textcolor{red}{\textbf{0.0282}}  & \textcolor{red}{\textbf{33.25}}\\
\bottomrule
\end{tabular}
}
\end{center}
\end{table*}

\begin{figure*}[t]
    \centering
  { 
      \includegraphics[width=0.98\linewidth]{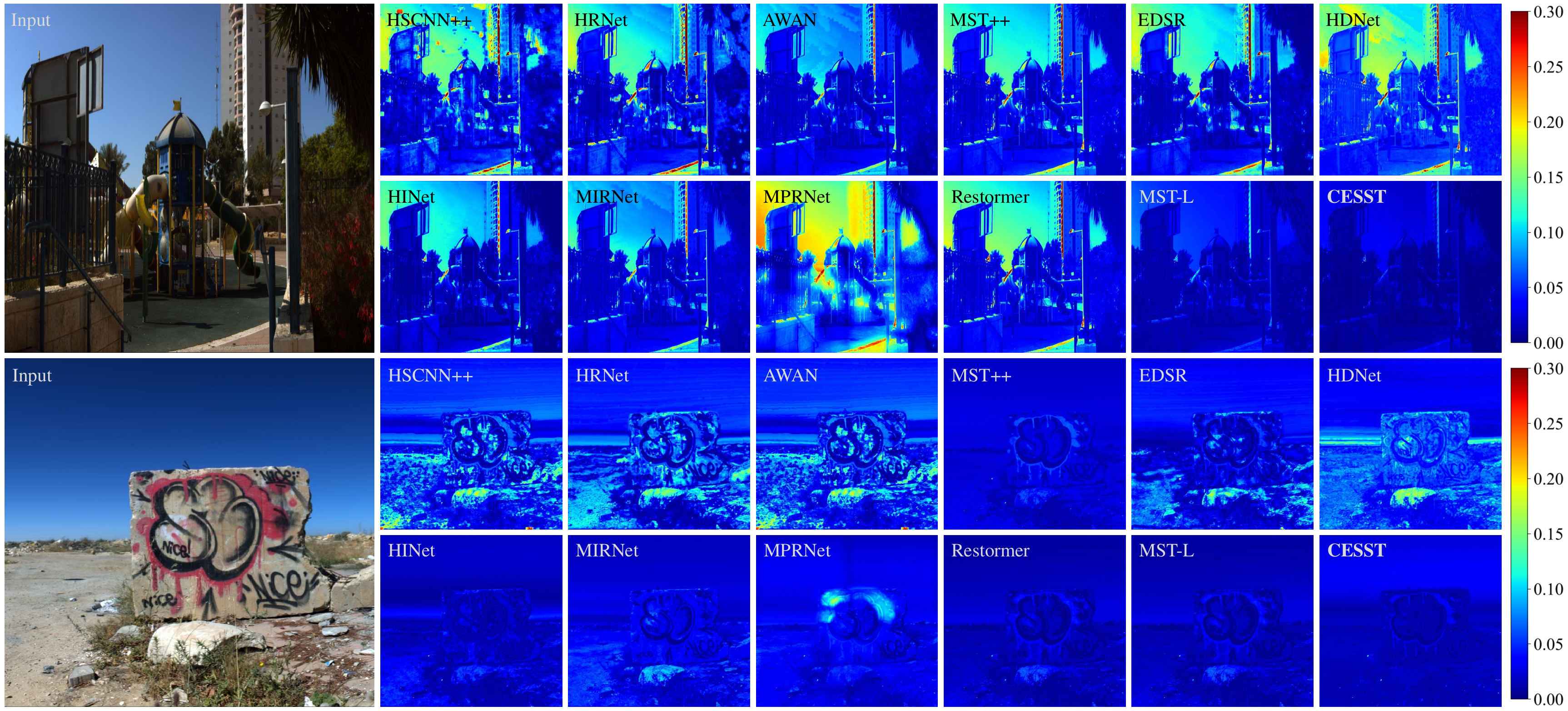}}
  \caption{The reconstruction error images of two images chosen from the validation set of ICVL dataset. The error images are the heat maps of the root mean square error (RMSE) (along spectral direction) between ground truths and reconstructed HSIs.} 
    \label{visual_compar_2} 
\end{figure*} 

\section{Experiments and Analysis}
\label{sec: experiment}

\subsection{Experimental Settings}
\textbf{Datasets.}
We adopt two datasets:  NTIRE2022 HSI dataset \cite{arad2022ntire} and ICVL HSI dataset \cite{Arad2016SparseRO}, to evaluate the performance of our CESST. In NTIRE2022 HSI dataset, there are 950 available RGB-HSI pairs, including 900 for training and 50 for validation. 
All the HSIs are captured at $482\times512$ spatial resolution over 31 channels from 400nm to 700nm. 
Besides, ICVL dataset contains 201 high-resolution HSIs. Considering that it does not provide aligned RGB images, we adopt the method proposed by Magnusson\textit{et al.} \cite{magnusson2020creating} to produce corresponding RGB images. Since it contains 18 images within different resolutions, we only use the left 183 image pairs (147 pairs for training and 36 pairs for testing).

\textbf{Implementation Details.}
We implement our CESST with Pytorch. All the models are trained with Adam \cite{kingma2014adam} optimizer ($\beta_1 = 0.9$ and $\beta_2 = 0.999$) for 300 epochs. The learning rate is initialized as $0.0002$, and the Cosine Annealing scheme is adopted.
During the training phase, RGB-HSI pairs are first cropped into $128\times128$ and the input RGB images are linearly rescaled to $[0, 1]$. We employ random rotation and flipping to augment training data.
The whole training time of the proposed CESST is about 40 hours with a single NVIDIA Ampere A100-40G.
All the RGB images are also rescaled to $[0, 1]$ during the validation procedure. Our CESST takes 0.141s per image (size of $482\times512\times3$) for HSI reconstruction. 

\textbf{Evaluation Metrics.}
We employ mean relative absolute error(MRAE), root mean square error(RMSE), Peak Signal-to-Noise Ratio(PSNR), error relative global dimensionless synthesis(ERGAS), and spectral angle mapper(SAM) as metrics to evaluate HSI reconstruction methods.

\subsection{Quantitative Results}
\label{quantity}
We compared our CESST with both HSI reconstruction and image restoration methods, including four RGB-HSI reconstruction methods: HSCNN+ \cite{shi2018hscnn+} (winner of NTIRE2018 HSI challenge), HRNet \cite{zhao2020hierarchical}, AWAN \cite{li2020adaptive} (winner of NTIRE2020 HSI challenge), MST++ \cite{cai2022mst++} (winner of NTIRE2022 HSI challenge); two compressive HSI recovery methods: HDNet \cite{hu2022hdnet} and MST-L \cite{cai2022mask}; five image restoration methods: EDSR \cite{lim2017enhanced}, HINet \cite{chen2021hinet}, MIRNet \cite{zamir2020learning}, Restormer \cite{zamir2022restormer}, and MPRNet \cite{zamir2021multi}. For fair comparisons, all the methods were retrained and tested with the same settings as MST++ \cite{cai2022mst++}. 
As shown in Table. \ref{quantitative_2022_1}, it can be observed that our method obtains the best results of all five metrics while costing the least Params on NTIRE2022 HSI dataset. 
To more intuitively illustrate the competitiveness of our method, we provide the PSNR-Params comparisons in Fig. \ref{performance}, including both HSI reconstruction methods and image restoration methods. The horizontal axis is the Params (memory cost), and the vertical axis is the PSNR (performance). As can be seen, our method takes up the upper-left corner, indicating the best efficiency.
Note that although the FLOPs of our model are larger than MST++ \cite{cai2022mst++}, benefited from our parallel calculation design, \textit{e.g.}, the three parallel branches of feature extraction, the parallel spatial-MSA and spectral-MSA of SSAB, our model achieves comparable inference time compared with MST++ on the same GPU.

\subsection{Qualitative Results}

\begin{table*}
    \vspace{-6pt}
  \caption{Ablation studies of break-down ablation and fusion scheme comparison.}
  \vspace{-6pt}
  \begin{subtable}{0.68\linewidth}
    \caption{Break-down ablation study of different pipelines.}
    \centering
    \label{pipeline comparison} 
    \resizebox{1\columnwidth}{!}{
    \begin{tabular}{lcccccccc}
    \toprule
    Baseline & Independent modeling & SSAB & SFAM & Params(M) & FLOPs(G) & MRAE & RMSE & PSNR \\ 
    \toprule
    \checkmark &  &  &  & 0.92 & 54.24 & 0.3207 & 0.0465 & 28.14  \\
    \checkmark & \checkmark &  &  & 1.03 & 74.75 & 0.2514 & 0.0351 & 31.83 \\
    \checkmark & \checkmark & \checkmark  & & 1.37 & 81.38 &  0.1917 & 0.0244 & 33.71 \\
    \checkmark & \checkmark & \checkmark & \checkmark  & 1.54 & 90.18 & \textcolor{red}{\textbf{0.1497}} & \textcolor{red}{\textbf{0.0225}}  & \textcolor{red}{\textbf{35.14}}  \\
    \bottomrule
    \end{tabular}
    }
  \end{subtable}
  \hfill
  \begin{subtable}{0.3\linewidth}
    \begin{center}
    \caption{Ablations of different fusion schemes.}
    \centering
    \label{fusion compare} 
    \resizebox{1\columnwidth}{!}{
    \begin{tabular}{lcccc}
    \toprule
    Method   &  MRAE & RMSE & PSNR \\ 
    \toprule
    \textit{Baseline (concatenation)}~       & 0.1917 & 0.0244 & 33.71 \\
    \textit{SFT \cite{wang2018recovering}}~ & 0.1652 & 0.0253 & 34.22 \\
    \textit{EFF \cite{hu2022hdnet}}~             & 0.1611 & 0.0242 & 34.73 \\
    \textit{SFAM (Ours)}~             & \textcolor{red}{\textbf{0.1497}} & \textcolor{red}{\textbf{0.0225}}  & \textcolor{red}{\textbf{35.14}} \\
    \bottomrule
    \end{tabular}
    }
    \end{center}
  \end{subtable}
  \label{table: ablations}
\end{table*}

Fig. \ref{visual_compar_3} visualizes the 30th band and the reconstruction error images of an HSI chosen from the validation set of NTIRE2022 HSI dataset \cite{arad2022ntire}, which corresponds to Fig. \ref{fig:short-a}.
Existing HSI reconstruction methods fail to generate chromatic-consistent content and artifact-free results, especially for the high-frequency components (\textit{}, the sky area).
In contrast, our method is capable of recovering more precise texture information and better pixel-level quality over other methods. This is because we treat each channel of input RGB images as a unique feature and model them into high dimensional space respectively, rather than equally modeling them and studying correlations in the high dimensional space directly (\textit{i.e.}, this is why our method perform better in the long-range spectral bands, as shown in \ref{fig:short-b}).
Besides benefiting from the independent modeling and combinatorial excavation mechanism, as well as our spatio-spectral attention mechanism, both spectral self-similarities and spatial details, can be efficiently explored and well fused to generate fine-grained pixel-level predictions. Meanwhile, the intensity of our results is closest to ground truths compared with existing methods.

In addition, to more comprehensively evaluate the visual quality of our method, we further provide more visual comparisons of error maps on ICVL dataset. As shown in Fig. \ref{visual_compar_2}, almost all existing methods fail to reconstruct the sky; for the second scene, all existing methods fail to reconstruct the foreground objects. In contrast, 
our method shows strong capabilities in recovering texture details, and preserving both spatial smoothness and harmonious tones of the homogenous regions.


\begin{figure}[t]
    \centering
  { 
      \includegraphics[width=0.95\linewidth]{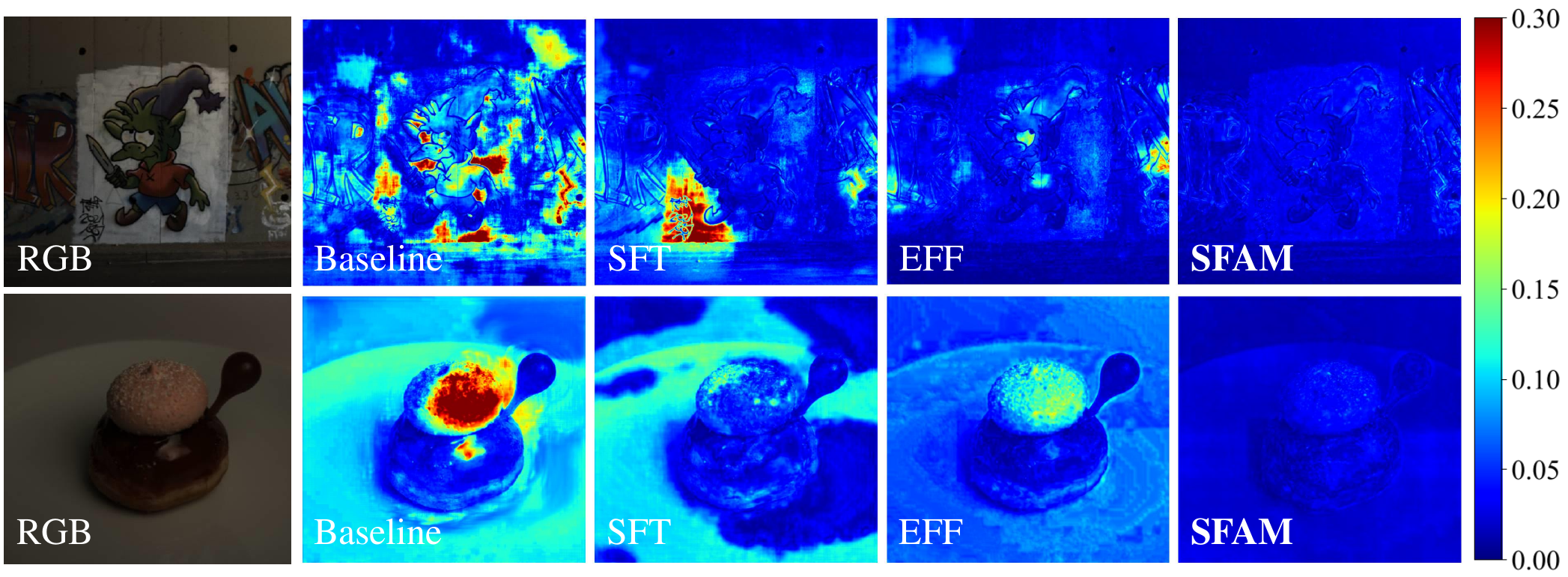}}
  \caption{The visual analysis of error images chosen from the validation set of NTIRE2022 HSI dataset. The error images are the heat maps of the root mean square error (RMSE) between the ground truth and the reconstructed HSI.}
    \label{visual_fusion_comaprison} 
\end{figure}

\subsection{Ablation Study}

In this section, we perform ablation studies to investigate the effectiveness of our proposed structure. 
The baseline model is derived by removing the independent modeling structure, including spatio-spectral attention block (SSAB), and spectrum-fusion attention module (SFAM) from CESST, and using the widely used ResNet \cite{he2016deep} block.

\textbf{Break-down Ablation.} 
To investigate the effect of each module, we first perform a break-down ablation and provide the quantitative results in Table \ref{pipeline comparison}. From the first and the second rows, we find that the independent modeling significantly improves the performance of the whole model, which yields a 3.69dB improvement in PSNR. When we successively apply both SSAB and SFAM, the reconstruction performance further achieves 1.88dB and 1.43dB improvement. These results demonstrate the effectiveness of our independent modeling, SSAB, and SFAM. 

\textbf{Fusion Scheme Comparison.} 
As the fusion module is one of the main contributions of this work, we further compare our SFAM with the other popular fusion schemes in Table \ref{fusion compare}, including concatenation-convolution (acts as the baseline in this scenario, which is the same settings as the third row in Table \ref{pipeline comparison}), SFT layer \cite{wang2018recovering}, and EFF layer \cite{hu2022hdnet}. As can be seen, our module gains 1.43dB, 0.92dB, and 0.41dB in PSNR, which verifies the effectiveness of our SFAM. 
In addition, we further provide visual analysis in Fig. \ref{visual_fusion_comaprison}, which shows that our SFAM is more capable of fusing fine-grained details, especially in the salient regions.

\textbf{MSA Comparison.}
To further validate the effectiveness of our proposed spatio-spectral attention mechanism, we compare our SSAB with different MSA variations, including spatial-MSA \cite{dosovitskiy2020image}, shifted-window MSA \cite{liu2021swin}, spectral-MSA \cite{cai2022mst++} and shuffle-window MSA (a variation of our original spatio-spectral MSA, which disables the spectral-MSA and keeps other settings consistent). We switch these modules directly in our framework. As shown in Table \ref{MSA comparison}, our spatio-spectral MSA performs best in RMSE and PSNR and is comparable with spectral-MSA in MRAE.

\begin{table}[t]
\small
\caption{Ablation study of different MSAs.}
\vspace{-6pt}
\centering
\label{MSA comparison} 
\resizebox{1\columnwidth}{!}{
\begin{tabular}{lcccc}
\toprule
MSA scheme   &  MRAE & RMSE & PSNR \\ 
\toprule
\textit{Spatial-MSA \cite{dosovitskiy2020image}}~  & 0.1614 & 0.0251 & 34.75 \\
\textit{Shifted-window MSA \cite{liu2021swin}}~     & 0.1783 & 0.0298 & 33.92 \\
\textit{Spectral-MSA \cite{zamir2022restormer}}~   & \textbf{0.1483} & 0.0241 & 34.96 \\
\textit{Shuffle-window MSA}~                      & 0.1639 & 0.0261 & 34.51 \\
\textit{\textbf{Spatio-spectral MSA} (ours)}~      & 0.1497 & \textbf{0.0225} & \textbf{35.14}\\
\bottomrule
\end{tabular}
}
\end{table}

\section{Conclusion}
\label{sec: conclusion}

In this paper, we have proposed a novel hyperspectral image reconstruction framework that excavates the unique and complementary information among the RGB input channels in a combinatorial manner for efficient embedding of spatio-spectral clues based on a Transformer structure: CESST. 
We have proposed a spatio-spectral attention module, and a spectrum-fusion attention module, which greatly facilitates the excavation and fusion of information at both semantically long-range levels and fine-grained pixel levels across all dimensions.
The effectiveness of each module has been validated by ablation studies.
Extensive visual comparisons and quantitative experiments have demonstrated that our proposed method achieves superior HSI reconstruction performance compared with SOTA methods.

\section*{Acknowledgments}
This research was supported by A*STAR C222812026.


\bibliography{aaai24}

\end{document}